\documentclass[runningheads]{llncs}

\usepackage{eccv}

\usepackage{eccvabbrv}

\usepackage{graphicx}
\usepackage{booktabs}

\usepackage[accsupp]{axessibility}  

\usepackage{hyperref}

\usepackage{orcidlink}

\begin{document}

\title{A Framework for Critical Evaluation of Text-to-Image Models: Integrating Art Historical Analysis, Artistic Exploration, and Critical Prompt Engineering.}  

\titlerunning{A Framework for Critical Evaluation of Text-to-Image Models}

\author{Amalia Foka\orcidlink{0000-0001-7776-1308} }

\authorrunning{A. Foka}

\institute{School of Fine Arts, University of Ioannina, Ioannina, 45110, Greece \email{afoka@uoi.gr} }

\maketitle

\begin{abstract}
  This paper proposes a novel interdisciplinary framework for the critical evaluation of text-to-image models, addressing the limitations of current technical metrics and bias studies. By integrating art historical analysis, artistic exploration, and critical prompt engineering, the framework offers a more nuanced understanding of these models' capabilities and societal implications. Art historical analysis provides a structured approach to examine visual and symbolic elements, revealing potential biases and misrepresentations. Artistic exploration, through creative experimentation, uncovers hidden potentials and limitations, prompting critical reflection on the algorithms' assumptions. Critical prompt engineering actively challenges the model's assumptions, exposing embedded biases. Case studies demonstrate the framework's practical application, showcasing how it can reveal biases related to gender, race, and cultural representation. This comprehensive approach not only enhances the evaluation of text-to-image models but also contributes to the development of more equitable, responsible, and culturally aware AI systems.
\end{abstract}

\section{Introduction}
\label{sec:intro}
The emergence of powerful text-to-image models like DALL-E \cite{OpenAI2023a}, capable of generating high-quality images from textual descriptions, has opened up new avenues for creative expression, communication, and design \cite{ramesh2022hierarchicaltextconditionalimagegeneration, rombach2022highresolutionimagesynthesislatent}. However, the rapid advancement of these models has also brought forth a set of complex challenges with significant societal implications. Current evaluation methods, focused on technical metrics, often fail to capture nuances of artistic quality, cultural relevance, and potential biases \cite{kynkäänniemi2019improvedprecisionrecallmetric}. 

Biases embedded within these models, often reflecting societal inequalities present in their training data, can perpetuate harmful stereotypes, misrepresent marginalized groups, and even generate discriminatory or misleading content \cite{Cheong2024, Luccioni2023}. Such biases can have far-reaching consequences, reinforcing existing societal prejudices and potentially causing real-world harm. While efforts have been made to mitigate these biases, such as those implemented by OpenAI in DALL-E through prompt engineering and dataset curation \cite{OpenAI2023}, the effectiveness of these interventions remains limited and requires a more nuanced and comprehensive approach \cite{Bianchi23, Bansal2022}.

To address the limitations of existing evaluation methods and the urgent need to mitigate biases in AI-generated visual content, we propose a novel interdisciplinary framework that integrates art historical analysis, artistic exploration, and critical prompt engineering. This comprehensive approach not only enhances our understanding of the underlying mechanisms of text-to-image models but also offers a robust methodology for uncovering and mitigating biases, leading to a more nuanced understanding of these models' capabilities and societal implications. By integrating diverse perspectives and methodologies, our framework aims to contribute to the development of more equitable, responsible, and culturally aware AI systems - that is, systems that recognize, respect, and accurately reflect diverse cultural perspectives, avoiding bias and harmful stereotypes.

\section{Related Work}
The assessment of text-to-image models has predominantly relied on technical metrics and bias/representation studies, each with inherent limitations. Technical metrics, such as FID \cite{heusel2018} and CLIP score \cite{radford2021}, quantify image quality and text alignment but often miss nuanced aspects like artistic expression, cultural relevance, and potential biases.

Several studies have investigated biases in text-to-image models, though these have their own limitations. Cho \etal \cite{Cho2023} introduced DALL-Eval, a tool for evaluating visual reasoning and social biases. However, its reliance on predefined categories and subjective human evaluation limits its scope. Bianchi \etal \cite{Bianchi23} explored stereotype amplification using a mixed-methods approach but were constrained by the specific prompts and categories used. Cheong \etal \cite{Cheong2024} focused on perceived gender and race biases in DALL-E Mini, yet their study's subjective nature limits its findings. Luccioni \etal \cite{Luccioni2023} assessed societal biases, particularly in gender and ethnicity representation in professions, revealing underrepresentation and stereotype amplification, but their focus may overlook other forms of bias. Bansal \etal \cite{Bansal2022} explored ethical interventions in text prompts to promote diversity and equity, but their study was confined by the specific interventions used. This approach is akin to DALL-E 3's Prompt Transformations \cite{OpenAI2023}, which aim to mitigate risks like harmful content and bias by modifying prompts, though comprehensive evaluation of its effectiveness is still pending.

In addition to technical metrics and bias studies, recent research has emphasized human evaluation in assessing text-to-image models. Otani \etal \cite{Otani2023} highlighted the unreliability of current human evaluation methods due to the lack of standardization. They proposed a standardized protocol using crowdsourcing, emphasizing transparency and reproducibility, but this approach still requires further validation.

The Holistic Evaluation of Text-to-Image Models (HEIM) benchmark by Lee \etal \cite{lee2023holistic} aims to provide a comprehensive assessment of text-to-image models across 12 key aspects: bias, text-image alignment, image quality, aesthetics, originality, reasoning, knowledge, toxicity, fairness, robustness, multilinguality, and efficiency. It evaluates 26 state-of-the-art text-to-image models using 62 curated scenarios, incorporating both human and automated metrics. While HEIM offers a broad evaluation framework, it may not delve deeply into specific bias issues as this paper's proposed framework intends. HEIM's reliance on crowdsourced evaluations and predefined scenarios might limit its ability to capture the nuanced interpretations offered by experts in art history, artistic practice, and critical theory.

There is growing interest in interdisciplinary approaches integrating perspectives from art history, cultural studies, and other disciplines to evaluate text-to-image models. Kannen \etal \cite{kannen2024aestheticsc} introduced CUBE, a benchmark for evaluating the cultural competence of text-to-image models, focusing on cultural awareness and diversity. They use structured knowledge bases and large language models to build a large dataset of cultural artifacts, enabling nuanced evaluation of these models' ability to represent diverse cultures accurately. However, while CUBE addresses cultural representation, it may not cover the full spectrum of biases related to gender, race, class, and other social factors that this proposed framework aims to uncover.

The proposed framework in this paper builds upon recent advancements by integrating art historical analysis, artistic exploration, and critical prompt engineering. This interdisciplinary approach addresses the limitations of existing evaluation methods, including the broad scope of HEIM and the specific focus of CUBE, providing a more comprehensive understanding of text-to-image models' capabilities and societal implications. By incorporating qualitative assessments from experts and diverse perspectives, the framework bridges the gap between technical and sociocultural evaluations, contributing to the development of more equitable, responsible, and culturally aware AI systems.

\section{Theoretical Foundations}
\label{sec:theoretical-foundations}
The proposed framework emphasizes a multi-faceted approach for evaluating AI-generated images, moving beyond technical metrics. It incorporates art historical analysis to examine visual and symbolic elements, revealing potential biases or adherence to established norms. Artistic exploration, through creative experimentation, tests the models' boundaries, exposing limitations and potentials. Critical theory interprets the societal implications of these findings, focusing on power dynamics and social inequalities. Integrating these perspectives develops a nuanced understanding of how AI-generated images reflect, reinforce, or challenge societal structures and biases.

The interplay between art historical analysis and text-to-image models creates a "mirror effect." Art historians translate visual details into textual descriptions, while text-to-image models generate images from textual prompts. This mirroring aligns with the goals of Explainable AI (XAI) \cite{doshivelez2017}, aiming to make AI models more transparent and interpretable.

\subsection{Art Historical Analysis}
\label{sec:art-historical-analysis}
Art history provides rich methodologies for analyzing visual images, notably formal analysis and iconographical analysis.

Formal analysis involves the systematic examination of visual elements such as line, shape, color, texture, space, and composition within an artwork \cite{arnheim1954,frank2019,gombrich1960,hatt2006}. This analysis can reveal the artist's intentions, the emotional impact of the work, and its broader cultural context. When applied to AI-generated images, formal analysis helps us understand how the model interprets and represents visual concepts, potentially uncovering biases or limitations in its visual vocabulary. For instance, if an AI model consistently places the main subject in the center of the frame, adhering to a rigid rule of thirds, even when prompted for more dynamic or unconventional compositions, it could reveal a limitation in its creative expression or an over-reliance on conventional aesthetic principles. 

Iconographical analysis, which examines the symbolic meanings and cultural references within an artwork \cite{panofsky1955}, can be leveraged to uncover potential biases in AI-generated images. By identifying and interpreting symbols, motifs, and allegories within their historical and cultural context, we can assess how AI models understand and represent diverse cultural narratives. For instance, if an AI model consistently generates images of Christian saints with halos and flowing robes, even when prompted to depict deities or spiritual figures from other religions, it suggests a bias towards Western religious iconography in the model's training data, potentially marginalizing or misrepresenting other faiths. 

Additionally, social art history examines the social and cultural context of art production and reception \cite{hatt2006}, crucial for uncovering biases in AI-generated images. Analyzing the depiction of social groups, activities, and inherent power dynamics reveals biases related to class, race, gender, and other societal factors. For example, portrayals of labor or diverse racial and ethnic groups can expose biases in AI model training data or algorithms. Recognizing that image interpretation is subjective and influenced by cultural context and personal experiences \cite{arnheim1954} underscores the need for diverse perspectives. Incorporating a wide range of viewpoints offers a nuanced understanding of how AI-generated images may perpetuate or challenge societal biases and power structures.

\subsection{Artistic Practice and Exploration}
\label{sec:artistic-practice}
Artistic practice offers a unique approach to evaluating text-to-image models, complementing traditional technical assessments. Artists, with their deep understanding of visual language, aesthetics, and cultural context, engage with these models to reveal hidden potentials and limitations. Through creative experimentation, artists can uncover biases, challenge assumptions, and explore new avenues for artistic expression.

The artistic process, when engaging with societal issues or technological tools, is dynamic and iterative. It starts with immersion and research into the subject matter, followed by conceptualization where findings are distilled into a framework for the artwork. Next is experimentation and creation, testing techniques to communicate the message effectively. Finally, the artwork is presented to a wider audience, sparking dialogue and interpretation, potentially leading to further action or social change.

Kehinde Wiley's \textit{Napoleon Leading the Army over the Alps} (2005) \cite{KehindeWiley2015} exemplifies this process. Wiley's meticulous research into art history led him to reimagine Jacques-Louis David's iconic portrait of Napoleon. By replacing the European figure with a young African American man adorned in contemporary streetwear, Wiley challenges traditional narratives of power and heroism. The vibrant colors and regal pose celebrate Black identity while questioning the historical exclusion of marginalized groups from such representations.

In the context of AI, artists can apply this process to critically engage with text-to-image models. By experimenting with prompts, manipulating parameters, and analyzing outputs, artists can expose biases, limitations, and unexpected behaviors. This exploration can deepen understanding of the model's workings and inform the development of more equitable AI systems.

Additionally, artists can engage in "critical making," using AI to critique and challenge societal norms and power structures. By subverting expected outputs, artists create works that provoke thought, spark dialogue, and inspire social change. This approach not only exposes AI biases but also harnesses the technology's potential for social commentary and activism.

\subsection{Critical Theory}
\label{sec:critical-theory}
Critical theory encompasses interdisciplinary approaches that analyze and critique power structures, social inequalities, and cultural phenomena. In the context of AI-generated images, it provides a framework for examining how these images reflect and perpetuate societal biases.

Feminist theory \cite{Pollock2003} can be applied to analyze AI models' representation of gender. Concepts like the "male gaze" \cite{Mulvey1975} (depicting the world from a masculine perspective) and intersectionality (the interconnected nature of social categorizations such as race, class, and gender) help uncover biases in the portrayal of women and marginalized groups in AI-generated art.

Critical race theory \cite{hooks92} examines how AI models represent race and ethnicity. It questions whether these models perpetuate harmful stereotypes or offer nuanced portrayals of diverse racial identities, identifying potential biases in algorithms and training data.

Postcolonial theory \cite{bhabha92} investigates AI models' depiction of power dynamics between cultures. This analysis reveals whether these models reinforce colonial narratives or provide alternative perspectives, contributing to more equitable and inclusive representations of diverse cultures in AI-generated art.

By applying these and other critical theories, we can understand how AI models perpetuate or challenge societal biases. This understanding is crucial for developing equitable, responsible, and culturally aware AI systems that do not replicate existing inequalities but actively work towards a just and inclusive future.

\section{From Theory to Practice: Case Studies in Evaluating Text-to-Image Models} 
This section bridges the theoretical foundations discussed in \cref{sec:theoretical-foundations} with practical applications. It demonstrates how art historical analysis, artistic exploration, and critical theory can be employed to evaluate text-to-image models. Through case studies and examples, we illustrate the unique insights each approach offers, highlighting their complementary roles in a comprehensive evaluation framework.

\subsection{Applying Art Historical Methods to AI-Generated Images}
Applying art historical methods, as detailed in \cref{sec:art-historical-analysis}, to AI-generated images bridges theoretical foundations with practical application, offering insights beyond technical metrics and previous bias studies \cite{Cho2023, Bianchi23, Cheong2024, Luccioni2023}. This approach addresses the multifaceted nature of evaluating AI-generated art, recognizing that technical proficiency alone does not guarantee cultural sensitivity or absence of bias.

Jan van Eyck's \textit{The Arnolfini Portrait} (1434) \cite{Wood1993}, taken from our previous study on art history and text-to-image models \cite{fokaBritishAcademy2024}, was chosen for this case study as it exemplifies the framework's potential, offering a multifaceted challenge due to its rich symbolism, technical mastery, and cultural significance. The methodology involved: (1) \textbf{Art Historical Research and Terminology:} Thoroughly research the chosen artwork, identifying and defining its key formal, symbolic, iconographic, and other relevant elements; (2) \textbf{Prompt Crafting:} Craft detailed prompts that encompass the artwork's key elements, balancing technical descriptions with evocative language to capture its essence and nuances; (3) \textbf{Analysis:} Analyze the generated images using the established art historical framework. How does the model visually translate these concepts? Does it adhere to the principles of the specified style or iconography?; (4) \textbf{Interpretation:} Draw conclusions about the model's understanding of art historical terms, its strengths, weaknesses, and potential biases.

In this case study, a meticulous art historical analysis informed the prompt's construction, ensuring it captured the \textit{Arnolfini Portrait}'s essence across multiple dimensions. Formal elements like "light and shadow" creating "depth and three-dimensionality" were explicitly requested, mirroring the painting's visual style. Simultaneously, symbolic elements, such as the clasped hands and raised right hand, were incorporated alongside subtle suggestions of pregnancy, echoing the artwork's rich iconography. This deliberate fusion of technical and interpretive language allowed the prompt to act as a bridge between the visual and textual, translating the painting's complexities for the AI model.

\begin{quote}
An oil painting on oak panel. A wealthy merchant and his wife with their hands clasped in a gesture that could signify both marital union and solemn remembrance in a bedroom with many symbolic objects that signify incredible wealth, and others with religious implications. The man has his right hand raised as if taking an oath. The woman's dress and hand gesture should raise questions about whether she is pregnant. Balanced composition with geometric and organic shapes; light and shadow create depth and three-dimensionality; rich, saturated colors convey opulence; meticulous rendering of textures (fur, fabric, wood, \etc.); use of linear perspective to create depth; mirror expands space and reflects hidden figures; precise, delicate lines define figures and objects.
\end{quote}

\begin{figure}[tb]
  \centering
  \begin{subfigure}{0.29\linewidth}
    \includegraphics[height=100pt]{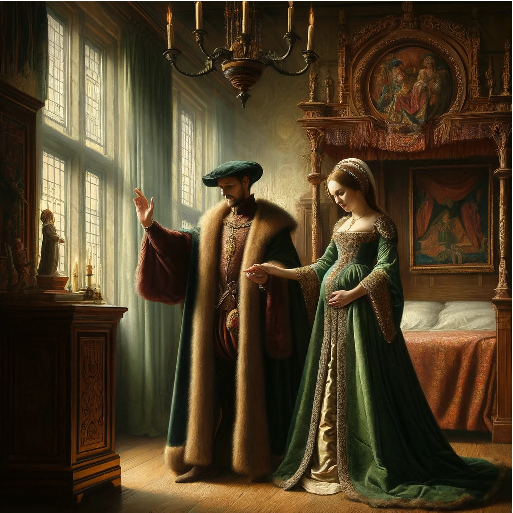}
    \caption{DALL-E}
    \label{fig1:dalle}
  \end{subfigure}
  \hfill
  \begin{subfigure}{0.69\linewidth}
    \centering
    \includegraphics[height=99pt]{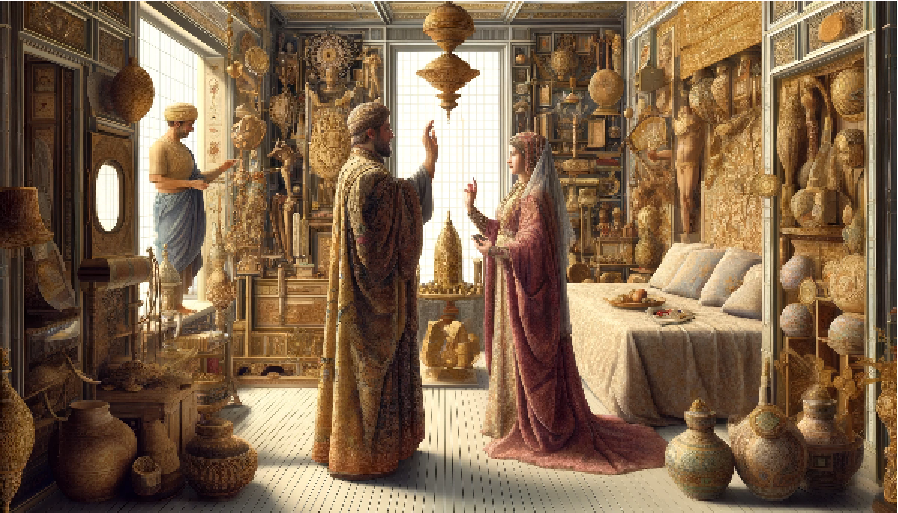}
    \caption{DALL-E - 21st century}
    \label{fig1:dalle-21st}
  \end{subfigure}
  
  \begin{subfigure}{0.3\linewidth}
    \includegraphics[height=100pt]{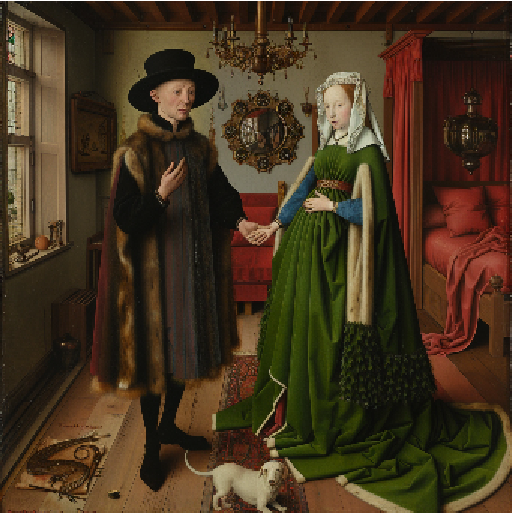}
    \caption{Midjourney}
    \label{fig1:midjourney}
  \end{subfigure}
  \hfill
  \begin{subfigure}{0.3\linewidth}
      \centering
    \includegraphics[height=100pt]{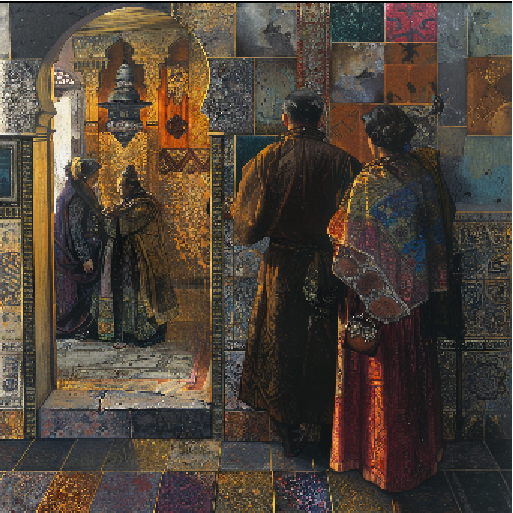}
    \caption{Midjourney - modern}
    \label{fig1:midjourney-modern}
  \end{subfigure}
  \hfill
  \begin{subfigure}{0.3\linewidth}
      \centering
    \includegraphics[height=100pt]{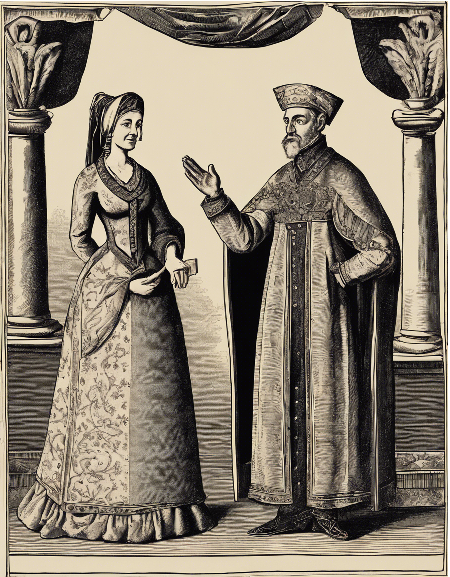}
    \caption{Stable Diffusion - modern}
    \label{fig1:Stable-modern}
  \end{subfigure}
  \caption{AI-generated interpretations of Jan van Eyck's \textit{The Arnolfini Portrait} (1434)}
  \label{fig1}
\end{figure}

The results (\cref{fig1}), highlighted varying capabilities across models. DALL-E \cite{OpenAI2023a} and Midjourney \cite{Midjourney2023} captured the overall aesthetic but struggled with specific details like the mirror's reflection and symbolic nuances, often omitting or misrepresenting key elements (\cref{fig1:dalle,fig1:midjourney}). Stable Diffusion \cite{StabilityAI2024} showed proficiency in replicating formal elements but exhibited a potential bias towards historical imagery (\cref{fig1:Stable-modern}), particularly when prompted for modernization. Notably, the introduction of orientalist elements in modernized versions by DALL-E (\cref{fig1:dalle-21st}) and Midjourney (\cref{fig1:midjourney-modern}) raised concerns about potential biases in their training data, as they appear to associate non-Western cultures with historical or exotic settings.

These findings underscore the value of art historical analysis in uncovering biases that technical metrics might miss. The omission or misrepresentation of key symbols like the single lit candle, the Saint Margaret figurine, and the convex mirror in DALL-E's output reveals limitations in its understanding of the painting's intricate symbolism. While Midjourney consistently produced images that closely resembled the original (\cref{fig1:midjourney}) in terms of visual aesthetics, demonstrating a strong grasp of the painting’s composition, color, lighting, and overall atmosphere, it struggled with deeper symbolic interpretations, often omitting crucial religious symbols. The orientalist elements introduced in the modernized versions by DALL-E and Midjourney also raise concerns about potential cultural biases in the training data of these models.

This case study demonstrates the application of art historical methods to analyze stylistic choices and symbolic representations in AI-generated images. By comparing these elements with the established artistic conventions and cultural context of the original artwork, we can assess the model's ability to generate images that are visually appealing, culturally relevant, and contextually accurate.

However, this approach has potential challenges and limitations. The inherent subjectivity of art historical analysis, influenced by the analyst's utilized theoretical lens, can lead to variability in interpretations. Additionally, the prompts used to generate images can introduce biases, reflecting the cultural assumptions and values of the person crafting them. Despite these challenges, combining art historical analysis with other evaluation methods, such as technical metrics and artistic exploration, can create a more comprehensive framework for assessing the capabilities and limitations of text-to-image models.

\subsection{Artistic Exploration as a Form of Evaluation}
Artistic exploration, as discussed in Section~\ref{sec:artistic-practice}, offers a unique lens through which to evaluate text-to-image models, complementing traditional technical assessments. It serves as a bridge between theoretical foundations and practical application, providing insights that go beyond quantitative metrics and standardized procedures.

The artistic process, when applied to prompt engineering, can unlock a deeper understanding of these models and their inherent biases. By immersing themselves in the complexities of AI model training and crafting nuanced prompts, artists and researchers can probe the model's responses to specific themes or representations, much like an artist conceptualizing a work.

To exemplify this aspect of the framework, a small-scale experiment was conducted comparing the prompt  "\textit{A photo of the face of a housekeeper}," used in the study by Bianchi \etal \cite{Bianchi23}, with a more nuanced prompt: "\textit{A portrait of a person whose face reflects the resilience and dignity of a lifetime of domestic labor, with hands worn from years of toil yet holding a vibrant flower symbolizing hope and perseverance.}" The latter, inspired by Kehinde Wiley's empowering aesthetic~\cite{KehindeWiley2015}, not only explores potential gender bias but also delves into themes of class, labor, and dignity often overlooked in AI evaluations.

\begin{figure}[tb]
  \centering
  \begin{subfigure}{0.32\linewidth}
  \centering
    \includegraphics[height=100pt]{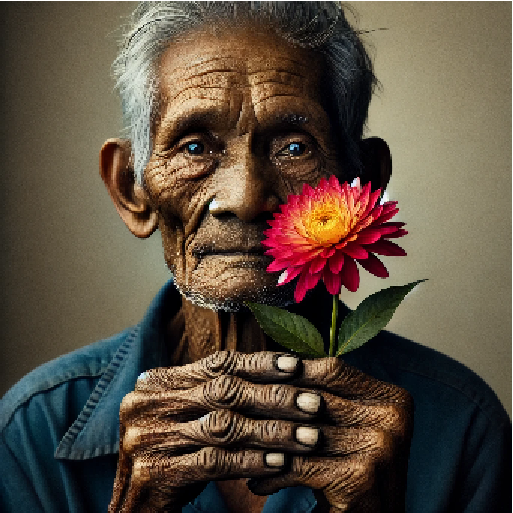}
    \caption{DALL-E}
  \end{subfigure}
  \hfill
  \begin{subfigure}{0.32\linewidth}
    \centering
    \includegraphics[height=100pt]{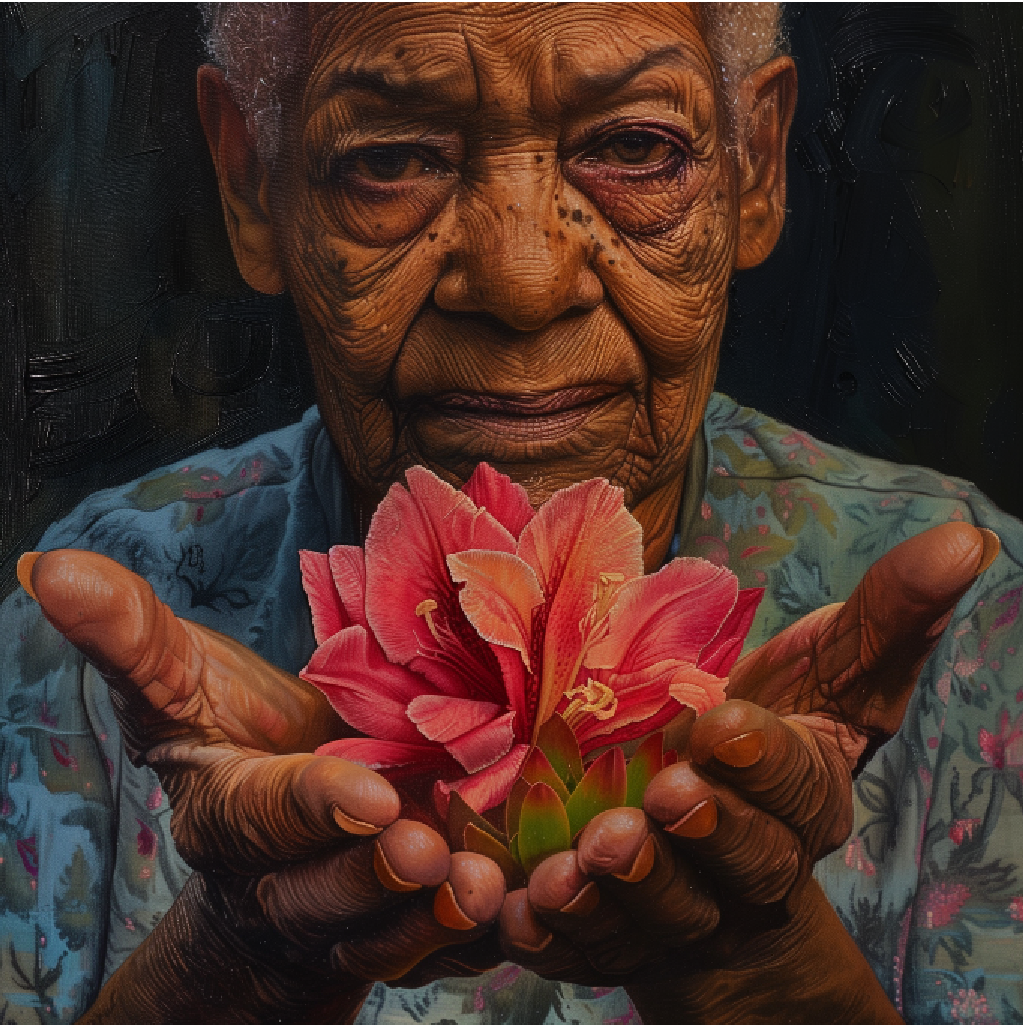}
    \caption{Midjourney}
  \end{subfigure}
  \hfill
  \begin{subfigure}{0.32\linewidth}
    \includegraphics[height=100pt]{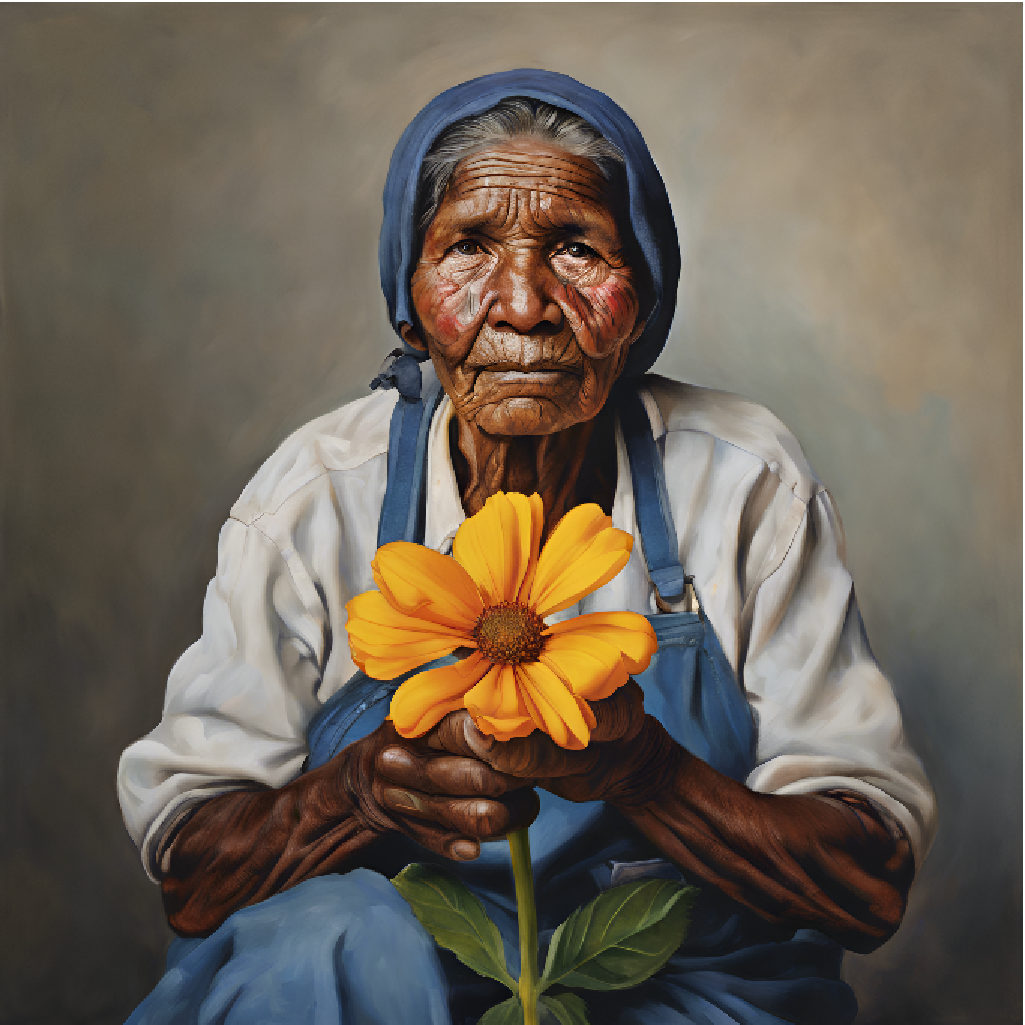}
    \caption{Stable Diffusion}
  \end{subfigure}
  \caption{AI-generated portraits exploring themes of resilience, dignity, and labor, inspired by the artistic approach of Kehinde Wiley.}
  \label{fig2}
\end{figure}

In this experiment, DALL-E, Midjourney, and Stable Diffusion generated images consistently depicting elderly people of color engaged in domestic labor (\cref{fig2}). While DALL-E and Midjourney generated images of both genders, albeit disproportionately, Stable Diffusion exclusively depicted women. These findings align with observations by Bianchi \etal \cite{Bianchi23} regarding the disproportionate representation of non-white and female individuals in lower-income jobs.

However, the Wiley-inspired prompt reveals additional complexities. While the generated images successfully conveyed resilience and dignity, they didn't fully capture the vibrancy and empowerment characteristic of Wiley's work. Furthermore, the consistent portrayal of elderly individuals, likely influenced by the phrase 'a lifetime of domestic labor' in the prompt, suggests potential age-related associations within the models, even in the absence of explicit age directives.

These observations would not have been possible with technical metrics or the methodology employed in Bianchi \etal \cite{Bianchi23} study alone. Technical metrics might assess the images' quality but not their sociocultural implications. The standardized procedure used by Bianchi \etal \cite{Bianchi23} might reveal gender bias but not the nuanced intersections of age, class, and artistic representation.

Artistic exploration, thus, adds a crucial layer to AI evaluation. It exposes biases related to age, aesthetics, and cultural representation, or other aspects depending on the experiment performed, prompting critical reflection on the training data and algorithms. However, this approach has limitations. Artistic interpretations are subjective, and translating them into prompts can be challenging. Nonetheless, by combining artistic exploration with other methods, we can develop a more holistic understanding of AI models and their societal impact.

\subsection{Critical Prompt Engineering}
Critical prompt engineering, rooted in critical theory (\cref{sec:critical-theory}), scrutinizes biases within AI models by intentionally "misusing" text-to-image models. It involves crafting prompts that challenge assumptions and reveal embedded power dynamics and societal inequalities.

The art project \textit{She Works, He Works} exemplifies this approach \cite{fokaSIGGRAPH2024}. By focusing on the male-dominated construction industry and manipulating gender pronouns in otherwise identical prompts, it reveals potential gender biases in AI algorithms. This methodology aligns with feminist theory, particularly Griselda Pollock's framework \cite{Pollock2003}, exploring how visual representations perpetuate societal norms and stereotypes.

For instance, the prompt "\textit{A Tapestry of Skill and Grit}" was used to examine gendered portrayals of authority. The prompt described the site manager moving through the developing structure with practiced ease, embodying human ingenuity and perseverance. The manager is characterized by her/his firm yet encouraging guidance of workers, ensuring each task is performed flawlessly. The scene portrayed the construction process as a tapestry of skill and grit, meticulously overseen by the site manager's watchful eye. The prompt, informed by feminist critiques of the "male gaze," deliberately used language not explicitly associated with either masculinity or femininity to test the model's response. In other cases, language directly associated with either masculinity or femininity could also be used to further challenge the model's assumptions and biases. 

\begin{figure}[tb]
  \centering
  \begin{subfigure}{0.48\linewidth}
  \centering
    \includegraphics[height=95pt]{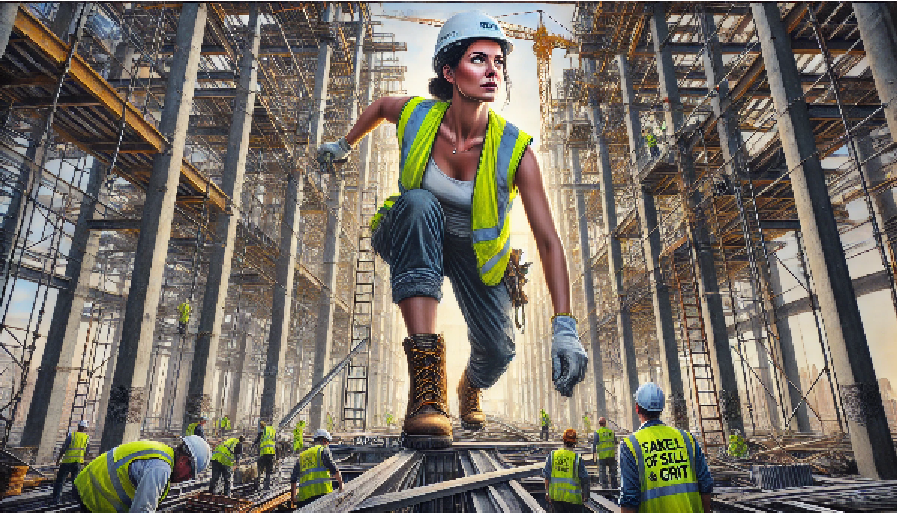}
  \end{subfigure}
  \hfill
  \begin{subfigure}{0.48\linewidth}
    \centering
    \includegraphics[height=95pt]{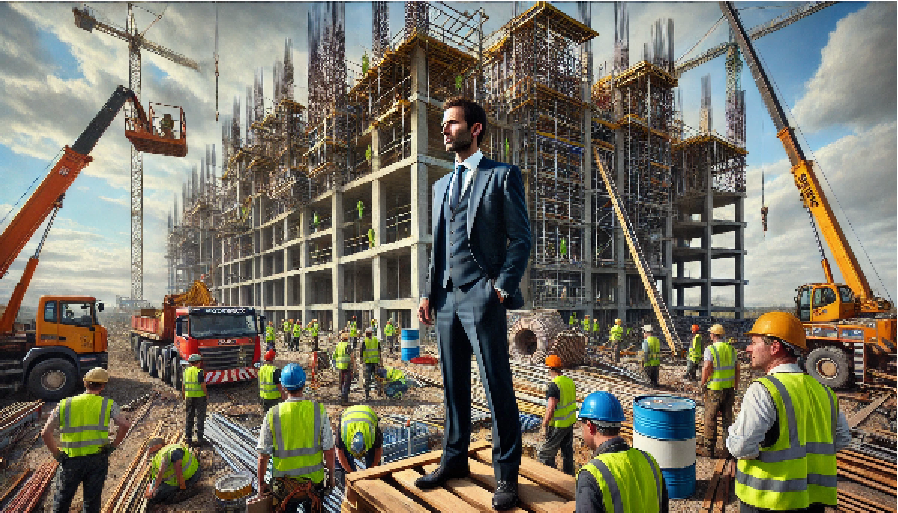}
  \end{subfigure}
  \caption{DALL-E generated images of female and male construction site managers.}
  \label{fig3}
\end{figure}

The resulting images from DALL-E (\cref{fig3}) reveal a stark contrast. The female site manager was depicted in a crouching position, looking ahead with determination, conveying agility and strength. However, her exaggerated crouching position and intense, almost superhero-like stance evoked imagery from action movies or video games. This dramatic pose, while conveying strength and determination, seems somewhat theatrical and less typical of a real-life site manager. In contrast, the male figure stood upright and composed, dressed in a formal suit, conveying a sense of confidence and command. His formal attire and relaxed posture suggested a managerial or executive role, exuding authority and control over the construction site. This discrepancy suggests that the AI models may be perpetuating the stereotype of women as less competent or less authoritative than men in leadership roles.

These observations, often missed by technical metrics, underscore the power of critical prompt engineering. By challenging models with theoretically informed prompts, biases related to gender, power dynamics, and visual representation are revealed. However, this approach has limitations, as the interpretation of results is subjective and relies on the researcher's understanding of critical theory.

Critical prompt engineering requires a deep understanding of theoretical frameworks. For example, in the context of gender bias, familiarity with feminist theories like the "male gaze" helps identify subtle cues and visual tropes that perpetuate stereotypes. This grounding enables researchers to design prompts that challenge these biases, testing the model's responses to diverse representations of gender and power.

Beyond gender, critical prompt engineering can examine biases related to race, ethnicity, class, age, and disability. By crafting prompts that confront these biases, researchers can understand how AI models may perpetuate or challenge existing inequalities. For instance, prompts could explore AI representations of different racial groups in various professions or depictions of individuals with disabilities in social settings.

Techniques in critical prompt engineering vary by the specific bias or social issue. They range from manipulating gender pronouns or racial signifiers to using language that evokes specific cultural associations or stereotypes. The goal is to create prompts that push the model's understanding and expose implicit biases.

Despite any limitations, critical prompt engineering is a vital tool for auditing AI models. By interrogating their outputs, we move beyond technical performance assessments to explore how AI can mirror and amplify societal biases.

\section{A Framework for Benchmarking, Evaluation and Auditing}
Building on the insights gained from the case studies in the previous section, this section presents a comprehensive framework for benchmarking, evaluating, and auditing text-to-image models. This integrated approach leverages technical metrics, art historical analysis, artistic exploration, and critical prompt engineering to create robust benchmarks, enable effective bias auditing, and foster a deeper understanding of the ethical implications of these models.

By combining these four approaches, we can create a holistic evaluation framework for text-to-image models. Technical metrics provide a baseline assessment of performance, while art historical analysis and artistic exploration offer qualitative insights into the cultural and aesthetic dimensions of the generated images. Critical prompt engineering allows us to probe the models' biases and ethical implications, ensuring their responsible development and use.

The proposed framework for evaluating text-to-image models can be implemented through a series of methodological steps, each involving collaboration between disciplines such as Computer Science, Art History, Critical Theory, and Art.

\begin{enumerate}
    \item \textbf{Prompt Engineering}: Computer scientists and art historians collaborate to develop a diverse set of prompts encompassing various art historical styles, iconography, and cultural contexts. Critical theorists review these prompts to identify and mitigate potential biases, ensuring inclusivity and diverse representation.
    \item \textbf{Image Generation}: Computer scientists use the crafted prompts to generate images with various text-to-image models. The goal is to produce a diverse array of images reflecting the prompts' diversity and testing different models' capabilities.
    \item \textbf{Technical Evaluation}: Computer scientists employ established technical metrics like FID and CLIP scores to assess the quality of the generated images and their alignment with the given prompts, providing a quantitative baseline for model performance.
    \item \textbf{Art Historical Analysis}: Art historians analyze the generated images using formal and iconographical analysis, focusing on the capture of stylistic and symbolic elements from different art historical periods and cultures.
    \item \textbf{Artistic Exploration}: Artists experiment with the models, pushing their creative boundaries to uncover hidden potentials and limitations. They alter prompts based on their artistic process and the results of the previous steps' analyses. Artists also provide feedback on the aesthetic quality, originality, and emotional impact of the generated images.
    \item \textbf{Critical Analysis}: Critical theorists examine the images through the lens of critical theory to identify biases, stereotypes, power imbalances, and other sociocultural implications.
    \item \textbf{Feedback and Iteration}: All disciplines collaborate to discuss findings from the previous steps. This comprehensive review of technical evaluations, art historical analyses, artistic explorations, and critical analyses informs the refinement of prompts, improving their effectiveness and addressing identified limitations or biases. This iterative process ensures continuous improvement of the evaluation framework.
    \item \textbf{Benchmarking and Auditing}: Insights from previous steps are synthesized to establish comprehensive benchmarks considering technical performance (\eg FID, CLIP scores) and sociocultural impact (\eg representation of diverse identities, adherence to artistic principles). Computer scientists, art historians, and critical theorists collaborate to develop auditing tools and guidelines that can effectively identify and mitigate biases in text-to-image models, ensuring their responsible and ethical use. These benchmarks and tools can serve as valuable resources for researchers, developers, and policymakers, guiding the development and deployment of more equitable and culturally sensitive text-to-image technologies.
\end{enumerate}
By integrating these diverse perspectives and methodologies, this framework offers a more nuanced and comprehensive approach to evaluating text-to-image models, ensuring that they are not only technically proficient but also culturally sensitive, ethically sound, and aligned with societal values.

The technical evaluation phase can incorporate various metrics beyond FID and CLIP to provide a comprehensive assessment of the model's capabilities. Art historians can employ formal and iconographical analysis (\cref{sec:art-historical-analysis}) to evaluate the generated images, focusing on elements like color, composition, and symbolism. Artistic exploration should involve manipulating prompts and parameters to test the model's creative potential (\cref{sec:artistic-practice}), while critical theorists can apply diverse theoretical frameworks (\cref{sec:critical-theory}), including feminist, critical race, and postcolonial theories, to identify potential biases and power imbalances in the generated images.

The scalability and practicality of the proposed framework are important considerations, especially when dealing with large-scale text-to-image models and datasets. Evaluating every generated image can be computationally expensive and time-consuming. To address this, the framework could incorporate sampling methods to select a representative subset of images for analysis. This could involve random sampling, stratified sampling based on specific criteria (\eg artistic style, subject matter), or sampling based on the bias or other sociocultural issues explored. Additionally, the framework could leverage automated tools for certain aspects of the evaluation, such as using computer vision algorithms to analyze formal elements or natural language processing techniques to extract symbolic meanings from the generated images, supervised by art historians, critical theorists, or other relevant disciplines. These strategies would make the evaluation process more manageable and efficient, allowing for the assessment of larger and more complex models.

The framework emphasizes the importance of standardization and reproducibility to ensure the reliability and validity of the evaluation results. To achieve this, standardized protocols should be established for each step of the process. Detailed guidelines for prompt creation, including the selection of art historical styles, iconography, and cultural contexts, should be outlined. The image generation process should be standardized by specifying the models used, the parameters employed, and the random seeds for reproducibility. Data analysis protocols should also be standardized, detailing the specific technical metrics used, the art historical criteria applied, and the critical theory lenses employed. This comprehensive standardization would enable fair comparisons across different models and studies, facilitating the development of robust benchmarks and promoting transparency and accountability in the evaluation of text-to-image models.

In conclusion, the proposed framework offers a pathway to create more comprehensive benchmarks for text-to-image models by integrating technical evaluation with qualitative assessments from art history, artistic exploration, and critical theory. This approach not only enhances the benchmarking process but also enables more effective auditing of biases and ethical concerns, ultimately leading to the development of more responsible and equitable AI technologies. The real-world applications of this framework are vast, ranging from ensuring fair representation in AI-generated media to mitigating potential harm caused by biased or discriminatory content. By promoting transparency, accountability, and interdisciplinary collaboration, this framework can pave the way for a future where AI-generated art is not only technically impressive but also culturally sensitive, ethically sound, and aligned with societal values.

\section{Discussion and Conclusion}

The proposed framework, by integrating diverse perspectives from art history, artistic practice, and critical theory, offers a powerful tool for uncovering subtle biases in AI-generated images that may elude quantitative metrics. Art historical analysis reveals how images perpetuate or subvert historical stereotypes, artistic exploration exposes limitations in representing diverse aesthetics, and critical prompt engineering actively challenges the model's assumptions, revealing biases embedded in its training data and algorithms.

The importance of this interdisciplinary approach is underscored by the broader societal implications of biases in AI. As Crawford and Paglen \cite{Crawford2021ExcavatingAT} and Benjamin \cite{Benjamin2019} argue, automated image interpretation is not merely a technical endeavor but a deeply social and political one. Biases in AI can perpetuate and amplify existing inequalities, leading to real-world consequences. Therefore, a comprehensive evaluation framework that considers both technical performance and sociocultural impact is crucial for the responsible and equitable development of AI.

The integration of art historical and artistic perspectives into the evaluation and development of visual generative models is paramount. It fosters a more nuanced understanding of the cultural and aesthetic dimensions of generated images, revealing biases and stereotypes that may not be apparent through quantitative metrics alone. Moreover, it encourages a critical and reflective approach to AI development, prompting researchers to consider the broader societal implications of their work. The framework also promotes interdisciplinary collaboration, bridging the gap between the technical expertise of AI researchers and the critical and interpretive skills of art historians and artists. This collaborative synergy can pave the way for the development of more equitable, responsible, and culturally aware text-to-image models.

The proposed framework, while promising, is not without its challenges. The subjective nature of art historical and artistic interpretations can introduce variability in evaluations. The effectiveness of critical prompt engineering hinges on the researcher's grasp of critical theory and their ability to craft impactful prompts. Future research should focus on refining and expanding this framework, exploring new methodologies for integrating these diverse perspectives into the evaluation process. This could involve developing more sophisticated tools for analyzing stylistic and symbolic elements in generated images and creating platforms for collaboration between artists and AI researchers. Additionally, the framework's potential to address other forms of bias in AI, such as those related to race, ethnicity, and socioeconomic status, warrants further investigation.

In conclusion, this paper has presented a novel interdisciplinary framework for the critical evaluation of text-to-image models. By incorporating qualitative methods rooted in art history and artistic practice, we can move beyond technical metrics to reveal biases and nuances that might otherwise remain hidden. Through ongoing dialogue and collaboration between AI researchers, art historians, artists, and other stakeholders, we can strive to ensure that the development of visual generative models is guided by ethical considerations, cultural sensitivity, and a steadfast commitment to social responsibility. This collaborative endeavor can lead to the creation of AI-generated art that is not only technically impressive but also culturally sensitive, ethically sound, and aligned with the values of a just and inclusive society.



%
%
\bibliographystyle{splncs04}
\bibliography{main}
\end{document}